# Combining Feature and Example Pruning by Uncertainty Minimization

**Marc Sebban**
Dpt of SJE
French West Indies and Guiana University
97159 Pointe-à-Pitre Cedex

**Richard Nock**
Dept of Mathematics and CS
French West Indies and Guiana University
97159 Pointe-à-Pitre Cedex

## Abstract

We focus in this paper on dataset reduction techniques for use in $k$-nearest neighbor classification. In such a context, *feature* and *prototype* selections have always been independently treated by the standard storage reduction algorithms. While this certifying is theoretically justified by the fact that each sub-problem is NP-hard, we assume in this paper that a joint storage reduction is in fact more intuitive and can in practice provide better results than two independent processes. Moreover, it avoids a lot of distance calculations by progressively removing useless instances during the feature pruning. While standard selection algorithms often optimize the accuracy to discriminate the set of solutions, we use in this paper a criterion based on an uncertainty measure within a nearest-neighbor graph. This choice comes from recent results that have proven that accuracy is not always the suitable criterion to optimize. In our approach, a feature or an instance is removed if its deletion improves information of the graph. Numerous experiments are presented in this paper and a statistical analysis shows the relevance of our approach, and its tolerance in the presence of noise.

## 1 INTRODUCTION

With the development of modern databases, data reduction techniques are commonly used for solving machine learning problems. In the Knowledge Discovery in Databases field (KDD), the human comprehensibility of the model is as important as the predictive accuracy of decision trees. To address the issue of human comprehensibility and build smaller trees, data reduction techniques has been proposed to reduce the number of instances (John, 1995) or features (Cherkauer & Shavlik, 1996) used during induction.

For Instance-Based Learning algorithms (IBL), data reduction techniques are also crucial even if they do not pursue the same aim. While Nearest-Neighbor (NN) algorithms are known to be very efficient for solving classification tasks, this effectiveness is counterbalanced by large computational and storage requirements. The goal of reduction procedures consists then in reducing these computational and storage costs. In this paper, we only focus on these storage reduction algorithms for use in a NN classification system.

The selection problem has been treated during last decades according to two different ways to proceed: (i) the reduction algorithm is employed to reduce the size $n$ of the learning sample, called Prototype Selection (PS) (Aha, Kibler & Albert, 1991; Brodley & Friedl, 1996; Gates, 1972; Hart, 1968; Wilson & Martinez, 1997; Sebban & Nock, 2000), (ii) the algorithm tries to reduce the dimension $p$ of the representation space in which the $n$ instances are inserted, called Feature Selection (FS) (John, Kohavi & Pfleger, 1994; Koller & Sahami, 1996; Skalak, 1994; Sebban, 1999).

Surprisingly, as far as we know, except Skalak (1994) who independently presents FS and PS algorithms in a same paper, and Blum and Langley (1997) who propose independent surveys of standard methods, no approach has attempted to globally treat the reduction problem, and to control these two degrees of freedom in the same time. Yet, Blum and Langley (1997) underline the necessity to further theoretically analyze the ways in which instance selection can aid the feature selection process. Authors emphasize the need for studies designed to help understand and quantify the relationship that relates FS and PS. In this paper, we attempt to deal with the storage reduction taken as a whole. Conceptually, the task is not so difficult, because PS and FS algorithms share the same aim and often optimize the same criterion. Theoretically speaking, the problem is in fact much more difficult: it can be actually proven that FS and PS are both NP-hard problems (reductions from the Set Cover problem). To cope with this difficulty, PS or FS algorithms propose heuristics to limit the search through the solution space, optimizing an adapted criterion. In this paper, we propose to extend this strategy to a joint reduction algorithm because it is still NP-hard. The majority of reduction methods maximize the accuracy on the learning sample. While the accuracy optimization is certainly a way to limit further processing's errors on testing, recent works have proven that it is not always the suitable criterion to optimize. In Kearns and Mansour (1996), a formal proof is given that explains why Gini criterion and the entropy should be optimized instead of the accuracy when a top-down induction algorithm is used to grow a decision tree. In this paper, we propose to extend this principle to a joint data reduction. Our approach minimizes a global uncertainty (based on an entropy) within a $k$NN graph built on the learning set (see section 2 for definitions). In such a graph, a given $\omega$ instance is not only connected to its $k$NN, but also to instances that have $\omega$ in their neighborhood (called *associates* in Wilson & Martinez (1997)). The interest of this procedure consists in better assessing the field of action of each instance $\omega$ and then minimizing the risk of a wrongly elimination. We have already independently tested our uncertainty criterion in the PS (Sebban & Nock, 2000) and FS (Sebban, 1999) fields. We briefly recall in section 3 the main results obtained with these two approaches. Afterwards, we propose in section 4 to combine and extend their principles in one and a same mixed reduction algorithm. We give some arguments that justify the use of this joint procedure rather than two independent processes. Our mixed backward algorithm, called $(FS + PS)RCG$, alternates the deletion of an irrelevant feature, and the removal of irrelevant instances. A survey of feature relevance definitions is proposed in (Blum and Langley, 1997). With regard to prototypes, we cluster in this paper irrelevant instances in four main categories. We detail them (see figure 1) because the nature of each irrelevant instance will determine the stage of its deletion in our algorithm.

The *first* belong to regions in the feature space with very few elements (see instance 1 on figure 1). Even if most of these few points belong to the same class, their vote is statistically a poor estimator, and a little noise might affect dramatically their contribution. It is also common in statistical analyses to search and remove such points, in regression, parametric estimations, etc. The *second* belong to regions where votes can be assimilated as being randomized. Local densities are evenly distributed with respect to the overall class distributions, which makes such regions with no information. The *third* instances belong to the most concentrated areas between the border and the center of each class. They are not irrelevant by the risk they bring during the generalization, such as for the second. They constitute irrelevant instances in the sense that removing such examples would not lead to a misclassification of an unseen instance. The *last category* concerns mislabeled instances (instance 4 on the figure). Labeling error can come from variable measurement techniques, typing errors, etc. In statistics, mislabeled instances belong to a larger class called *outliers*, i.e. they do not follow the same model of the rest of the data. It includes not only erroneous data but also "surprising" data (John, 1995). A data reduction technique is called robust if it can withstand outliers in data. In such a context, we think that mislabeled instances must be eliminated prior to applying a given learning algorithm, even if a separate analysis could reveal information about special cases in which the model does not work. The third category must also be removed because it brings no relevant information. We are then essentially concerned with the progressive elimination of these instances during our mixed procedure. The remaining irrelevant instances will be removed later (when the representation space will be stabilized), because their status could evolve during the reduction of the representation space.

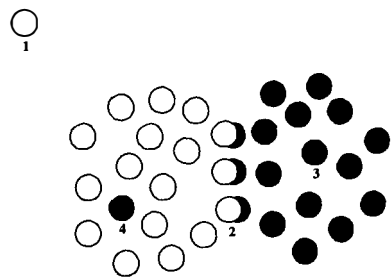

Figure 1: Four categories of irrelevant instances removed by a prototype selection algorithm

## 2 THEORETICAL FRAMEWORK

In this paper, we do not use the accuracy as criterion. Our performance measure is based on a quadratic entropy. The strategy consists in measuring information around each instance, and deducing a global uncertainty within the learning sample. The information is represented by the edges linked to each instance in the $k$NN unoriented graph. This way to proceed brings several advantages in comparison with the accuracy:

- contrary to the accuracy, our criterion does not depend on a *ad hoc* learning algorithm,

- with the accuracy, an instance is correctly classified or misclassified. The use of our criterion allows us to have a continuous certainty measure,

- we show below that our criterion has statistical properties which establish a theoretical framework for halting search,

- around a given instance $\omega$, we take into account not only the neighbors of $\omega$ but also instances which have $\omega$ in their neighborhood. Therefore, it allows to better assess the field of action of each instance.

**Definition 1** *The Quadratic Entropy is a function $QE$ from $[0,1]^c$ in $[0,1]$,*

$$QE : S_c \to [0,1]$$
$$(\gamma_1, .., \gamma_c) \to QE((\gamma_1, .., \gamma_c)) = \sum_{j=1}^{c} \gamma_j(1 - \gamma_j)$$

**Definition 2** *The neighborhood $N(\omega_i)$ in a $kNN$ graph of a given $\omega_i$ instance belonging to a sample $S$ is:*

$$N(\omega_i) = \{\omega_j \in S \ / \ \omega_i \text{ is linked by an edge to } \omega_j \text{ in the } kNN \text{ graph}\}$$

**Definition 3** *the local uncertainty $U_{loc}(\omega_i)$ for a given $\omega_i$ instance belonging to $S$ is defined as being:*

$$U_{loc}(\omega_i) = \sum_{j=1}^{c} \frac{n_{ij}}{n_{i.}}(1 - \frac{n_{ij}}{n_{i.}})$$

*where $n_{i.} = card\{N(\omega_i)\}$*

*and $n_{ij} = card\{\omega_l \in N(\omega_i) \ | \ Y(\omega_l) = y_j\}$ where $Y(\omega_l)$ describes the class of $\omega_l$ among $c$ classes.*

**Definition 4** *the total uncertainty $U_{tot}$ in the learning sample is defined as being:*

$$U_{tot} = \sum_{i=1}^{n} \frac{n_{i.}}{n_{..}} \sum_{j=1}^{c} \frac{n_{ij}}{n_{i.}}(1 - \frac{n_{ij}}{n_{i.}})$$

*where $n_{..} = \sum_{i=1}^{n} n_{i.} = 2card\{E\}$*

*where $E$ is the set of all the edges in the $kNN$ graph, and $n$ the number of instances.*

Light and Margolin (1971) show that the distribution of the relative quadratic entropy gain is a $\chi^2$ with $(n-1)(c-1)$ degrees of freedom. Rather than taking $U_{tot}$ as criterion, we define the following criterion:

**Definition 5** *The Relative Certainty Gain in a $kNN$ graph is defined as being:*

Table 1: Average accuracy on 20 databases; $p$ is the average size of the original sets, and $p^*$ is the size of the reduced space by $FSRCG$

| $p$ | $p^* = |E|$ | $FSRCG$ | Acc. | All |
|---|---|---|---|---|
| 15.7 | 8.4 | 76.69 | 75.05 | 73.71 |

$$RCG = \frac{U_0 - U_{tot}}{U_0}$$

where $U_0$ is the uncertainty computed directly from the *a priori* distribution.

$$U_0 = \sum_{j=1}^{c} \frac{n_j}{n}(1 - \frac{n_j}{n})$$

where $n_j = card\{\omega_i \ / \ Y(\omega_i) = y_j\}$, *i.e.* where the label of $\omega_i$ is the class $y_j$

Thus, $RCG$ is able to assess the effect of an instance or a feature deletion on the global uncertainty. These statistical properties constitute a rigorous framework and will provide an efficient criterion for halting search through the solution space.

## 3 INDEPENDENT FEATURE AND PROTOTYPE SELECTIONS

### 3.1 FSRCG

In this feature selection algorithm, proposed in Sebban (1999), we start without attribute (forward algorithm) and select the best feature that allows to reduce the global uncertainty $U_{tot}$. According to John, Kohavi and Pfleger (1994), this approach is a filter model, because it filters out irrelevant attributes before the induction process occurs (see also the following filter models: Koller & Sahami, 1996; Sebban, 1999). A filter method contrasts with wrapper models which use an induction algorithm (and then its accuracy) to assess the relevance of feature subsets (John, Kohavi and Pfleger, 1994; Skalak, 1994). The pseudocode of $FSRCG$ is presented in figure 2, where $>>$ means "statistically higher". Table 1 recalls a comparison between $FSRCG$ in a 1NN graph, a standard wrapper model optimizing the accuracy (Acc. in the table), and a model using all the attributes (All) (for details on experiments, see Sebban (1999)). The comparison on 20 benchmarks coming for the majority from the UCI repository was done using a 5-folds Cross-Validation procedure and applying a 1NN classifier.

A statistical analysis allows us to make the following remarks:

1. On average, *FSRCG* presents better results than

```
FSRCG ALGORITHM
RCG = 0  E = ∅; X = {X_1, X_2, ..., X_p}  Stop=false
Repeat
   For each X_i ∈ X do Compute RCG_i in E∪X_i
   Select X_min with RCG_min = Max{RCG_j}
   If RCG_min >> RCG then
      X = X - {X_min}
      E = E ∪ {X_min}
      RCG ← RCG_min
   else Stop:=true
Until Stop=true
Return the feature subset E
```

Figure 2: Pseudocode of $FSRCG$

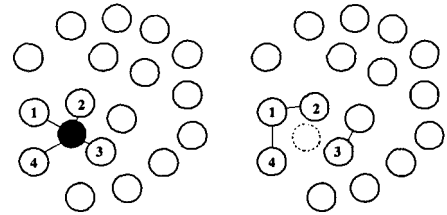

Figure 3: The deletion of an instance results in some local modifications of the neighborhood. Only instances (1,2,3 and 4) concerned by the deleted point (the black ball) are affected by these modifications.

a standard feature selection algorithm optimizing the accuracy. The mean gain of $FSRCG$ is about +1.6 (76.69 vs 75.05). Using a Student paired t test, we find that $FSRCG$ is statistically higher than **Acc.** with a critical risk near 5%, that is highly significant.

2. The advantage of $FSRCG$ is confirmed by analyzing the results of *All Attributes* (**All**). $FSRCG$ allows a better accuracy, on average +3.0, that is also highly significant. Using a Student paired t test, we find actually a critical risk near 0.6%, while the standard wrapper is better with a risk about 17%, that is less significant.

3. Globally, $FSRCG$ reduces the number of features (8.4 vs 15.7 on average) and then the storage requirements for a $k$NN classifier.

### 3.2 PSRCG

The high majority of PS algorithms uses the accuracy to assess the effect of an instance deletion (see Hart, 1968; Gates, 1972; Aha, Kibler & Albert, 1991; Skalak, 1994; Wilson & Martinez 1997). In Hart (1968), the author proposes a *Condensed NN Rule* (CNN) to find a *Consistent Subset*, *CS*, which correctly classifies all of the remaining points in the sample set. The *Reduced NN Rule* (RNN) proposed by Gates (1972) searches in Hart's *CS* for the minimal subset which correctly classifies all the learning instances. In Aha, Kibler and Albert (1991), the IB2 algorithm is quite similar to Hart's CNN rule, except it does not repeat the process after the first pass. Skalak (1994) proposes two algorithms to find sets of prototypes for NN classification. The first one is a Monte Carlo sampling algorithm, and the second applies random mutation hill climbing. Finally, in the algorithm $RT3$ of Wilson and Martinez (1997) (called *DROP3* in Wilson and Martinez (1998) and its extensions *DROP4* and *DROP5*), an instance $\omega_i$ is removed if its removal does not hurt the classification of the instances remaining in the sample set, notably instances that have $\omega_i$ in their neighborhood (called *associates*).

In our prototype selection algorithm, proposed in Sebban & Nock (2000), we insert our information criterion in a PS procedure. At each deletion, we have to compute again some local uncertainties, and analyze the effect of the removal on the remaining information (see figure 3). This update is relatively computationally unexpensive for the reason that only points in the neighborhood of the removed instance will be concerned by a modification. Updating a given $U_{loc}(\omega_j)$ after removing $\omega_i$ does not cost more than $O(n_{i.})$, a complexity which can be further decreased by the use of sophisticated data structures such as $k$-$d$ trees (Sproull, 1991). Our strategy consists in starting from the whole set of instances and searching for the worst instance which allows, once deleted, to improve the information. An instance is eliminated at the step $t$ if and only if two conditions are filled: (1) the Relative Certainty Gain after the deletion is better than before, i.e. $RCG_t >> RCG_{t-1}$ and (2) $RCG_t >> 0$.

Once this first procedure is finished (consisting in a way in deleting border instances), our algorithm executes a post-process consisting in removing the center of the remaining clusters and mislabeled instances. To do that, we increase the geometrical constraint by rising the neighborhood size ($k$+1), and remove instances that still have a local uncertainty $U_{loc}(\omega_i) = 0$. The pseudocode of $PSRCG$ is presented in figure 4. From table 2 which presents the summary of a wide comparison study (with $k = 5$) between the previous cited standard PS algorithms (see in Sebban and Nock, 2000 for more details), we can make the following remarks:

1. $CNN$ and $RNN$, which are known to be sensitive to noise fall much in accuracy after the in-

Table 2: Comparison between PSRCG and existing PS algorithms on 20 datasets; 5NN is inserted for comparison

| 5NN | CNN | | RNN | | PSRCG | | M. Carlo | RT3 | |
| --- | --- | --- | --- | --- | --- | --- | --- | --- | --- |
| Acc | size(%) | $Acc_{CNN}$ | size(%) | $Acc_{RNN}$ | size(%) | $Acc_{PSRCG}$ | $n_s=100$ | size(%) | $Acc_{RT3}$ |
| 75.7 | 44 | 72.4 | 41 | 71.9 | 44.9 | 75.9 | 74.6 | 12.6 | 73.7 |

PSRCG ALGORITHM
$t \leftarrow 0; N_p = |S|$
Build the kNN graph on the learning set
Compute $RCG_1$
$$RCG_1 = \frac{U_0 - U_{tot}}{U_0}$$
Repeat
  $t \leftarrow t+1$
  Select $\omega$ with the $\max(U_{loc}(\omega_j))$
  If 2 instances have the same $U_{loc}$
  select the example having the smallest
  number of neighbors
  Local modifications of the kNN graph
  Compute $RCG_{t+1}$ after removing $\omega$
  $N_p \leftarrow N_p - 1$
Until $(RCG_{t+1} < RCG_t)$ or $\text{not}(RCG_{t+1} >> 0)$
  Remove instances having a null uncertainty
with their $(k+1)$-NN
Return the prototype set with $N_p$ instances

Figure 4: Pseudocode of $PSRCG$

stance reduction (the difference is statistically significant).

2. The Monte Carlo method (Skalak) presents interesting results despite the constraint to provide in advance the number of samples and the number of prototypes (here $N_p$).

3. $RT3$ is suited to dramatically decrease the size of the learning set (12.6 % on average). In compensation, the accuracy in generalization is a little reduced (about -2.0).

4. $PSRCG$ requires a higher storage (44.9% on average), but interestingly, it allows here to slightly improve the accuracy (75.9% vs 75.7%) of the standard $k$NN classifier, even if a Student paired t test shows that the two methods are statistically indistinguishable. Nevertheless, it confirms that $PSRCG$ seems to be a good solution to select relevant prototypes and then reduce the memory requirements, while not compromising the generalization accuracy.

We have also tested the sensitivity of our algorithm to noise. A classical approach to deal with this problem consists in adding artificially some noise in the data set. This was done here by randomly changing the output class of 10% of the learning instances. We did not test the $CNN$ and $RNN$ algorithms on noisy samples, because they are known to be very sensitive to noise (Wilson & Martinez, 1998). Table 3 shows the average accuracy and storage requirements for $PSRCG$, $RT3$, and the basic $k$NN over the datasets already tested.

Table 3: Average accuracy and storage requirements when noise is inserted

| PS Algorithm | Noise-Free | Size % | Noisy | Size % |
| --- | --- | --- | --- | --- |
| kNN | 74.4 | 100.0 | 69.0 | 100.0 |
| PSRCG | 74.7 | 44.5 | 73.4 | 47.8 |
| RT3 | 72.5 | 11.0 | 69.7 | 10.5 |

$PSRCG$ has the highest accuracy (73.4%) of the three algorithms, while storage requirements are relatively controlled. It presents a good noise tolerance because it does not fall much in accuracy (-1.3 for 10% of noise), versus -2.8 for $RT3$ and -5.4 for the basic $k$NN.

## 4 COMBINING FEATURE AND PROTOTYPE SELECTION

### 4.1 PRESENTATION

The previous section has recalled that our information criterion seems to be efficient in a storage (feature or prototype) reduction algorithm. Since we optimize in both algorithms the same criterion, it is in the nature of things to combine the two approaches in one and the same selection algorithm. Nevertheless, this isolated argument is not very convincing and does not give information about the way to proceed. The following arguments will certainly bring a higher significance in favor of a combined selection, rather than two independent procedures.

- By alternatively reducing the number of features and instances, we reduce the algorithm complexity. The computation of a local uncertainty $U_{loc}(\omega)$ (useful for PS) is actually not very costly because it is done during the $RCG$ evaluation. Then, this additional PS treatment is not computationally expensive. On the other hand, this pro-gressive instance reduction will decrease a lot the future computations in the smaller representation space, avoiding numerous distance calculations on the removal examples. In total, a combined selection will be certainly less expensive than a feature selection followed by a only one global prototype selection.

- A progressive instance reduction during the feature pruning can also avoid to reach a local optimum during feature pruning. The interest of our algorithm is that it takes into account two criteria, one local ($U_{loc}(\omega)$), the other global ($RCG$). It consists in improving local information, without damaging the global uncertainty. The experiments will confirm this remark.

- Mixed approach attempts has been rare in this field, despite a common goal in PS and FS, (i.e. reducing the storage requirement) and obvious interactions (Blum and Langley, 1997).

Rather than roughly deleting a lot of instances after each stage of the feature reduction, we prefer to apply a moderate procedure which consists in sequentially removing only *mislabeled instances* and *center examples* during the mixed procedure. Actually, the status of these points will probably not evolve during the storage reduction. In our approach, we define a mislabeled instance as a $\omega$ example having in its $N(\omega)$ (including its associates) only points belonging to a different class of $\omega$, i.e.

$$\omega \text{ is mislabeled if and only if } Y(\omega) \neq Y(\omega_j),$$
$$\forall \omega_j \in N(\omega)$$

We can believe that few mislabeled instances in a $p$ dimension could become informative in a $(p-1)$-dimensional space.

Concerning instances that belong to the center of classes, their usefulness is not very important, and moreover they probably will not be disturbed by a feature reduction. Such instances have neighbors and associates belonging to their own class, i.e.

$$Y(\omega) = Y(\omega_j), \forall \omega_j \in N(\omega)$$

In conclusion, mislabeled or center instances have the common property to present a $U_{loc}(\omega) = 0$. We decide then to remove after each feature reduction all the instances that have in this new space a local uncertainty $U_{loc}(\omega) = 0$.

On the other hand, border points will be treated only when the feature space will be stabilized. Actually, boundaries of classes will certainly change during the feature reduction, and then border instances will evolve during the process. The pseudocode of our combined algorithm, called $(FS + PS)RCG$, is presented in figure 5.

(FS+PS)RCG ALGORITHM
$E = \{X_1, ..., X_p\}$; Compute $RCG$ in E; Stop=False
Repeat
    For each $X_i \in E$ Compute $RCG_i$ in $E - \{X_i\}$
    Select $X_{min}$ with $RCG_{min} = Min\{RCG_j\}$
    If $RCG_{min} >> RCG$ then
        $E = E - \{X_{min}\}$
        $RCG \leftarrow RCG_{min}$
        Remove mislabeled + center instances
        that have $U_{loc}(\omega) = 0$
    else Stop $\leftarrow$ true
Until Stop=true
Repeat
    Compute $RCG_0$ in $E$
    Delete $\omega$ that have a maximum $U_{loc}(\omega)$
    Compute $RCG_1$ after the deletion of $\omega$
Until ($RCG_1$<$RCG_0$) or not($RCG_1$>>0)

Figure 5: Pseudocode of $(FS + PS)RCG$

### 4.2 EXPERIMENTAL RESULTS AND COMPARISONS

The experiments presented in this section has been realized in order to bring to the fore the properties of $(FS + PS)RCG$ in the following situations:

1. *Ability of $(FS + PS)RCG$ to deal with global prototype and feature selection*

   To address this problem, we apply our algorithm on 17 benchmarks. We compare $(FS+PS)RCG$'s performances with a standard $k$NN ($k = 5$) classifier using all the features and all the instances. To limit the computational costs in this section due to a large comparison between strategies, we estimate the generalization accuracy only on a single validation sample (1/3 of the original instances). We can notice from results in the table 4 that despite a high storage reduction (about 50% for the number of instances on average, and 33% on features on average), $(FS+PS)RCG$ improves the generalization accuracy (77.3% versus 76.8%). Using a Student paired t test, we find that $(FS + PS)RCG$ is statistically higher than $k$NN with a risk near 26%.

2. *Ability of $(FS + PS)RCG$ to cope with noise in data*

   We add artificially some noise in the datasets, by randomly changing the output class of 10% of the learning instances. In the presence of noise, results of the table 4 show that $(FS + PS)RCG$ seems to be robust. Three arguments are justifying this claim: (i) while a standard $k$NN classifier gives an accuracy about 71.5% on noisy data, $(FS+PS)RCG$ clearly improves in the same context this accuracy (74.3%). Statistically speaking, $(FS+PS)RCG$ is significantly higher than a $k$NN with a risk near 2%; (ii) while a basic $k$NN classifier falls much in accuracy in the presence of noise (71.5% vs 76.8), $(FS + PS)RCG$ better controls the generalization accuracy (74.3% vs 77.3%); (iii) we note that the storage reduction of $(FS+PS)RCG$ is higher in the presence of noise than in a noise free situation (60% versus 50% on instances, and 36.4% versus 33% on features). This difference is explained by the necessity to remove more irrelevant information.

Table 4: Comparison between $(FS + PS)RCG$ and a $k$NN classifier in noise free and noisy situations

| Dataset | 5NN | | | (FS+PS)RCG | | | 5NN noisy | (FS+PS)RCG noisy | | |
|---|---|---|---|---|---|---|---|---|---|---|
| | Size | Dim | Accuracy | Size | Dim | Accuracy | Accuracy | Size | Dim | Accuracy |
| LED+17 (2 cl) | 300 | 24 | 66.5 | 191 | 16 | 75.0 | 60.0 | 80 | 16 | 66.5 |
| LED (2 cl) | 300 | 7 | 83.0 | 179 | 5 | 88.0 | 78.5 | 217 | 4 | 84.0 |
| Wh. House | 235 | 16 | 89.5 | 103 | 11 | 90.5 | 80.5 | 129 | 11 | 93.5 |
| Hepatitis | 100 | 19 | 70.9 | 68 | 7 | 69.1 | 65.5 | 41 | 9 | 70.9 |
| Horse Colic | 200 | 22 | 70.8 | 84 | 9 | 72.1 | 63.7 | 53 | 10 | 64.9 |
| Echocardio | 70 | 6 | 62.3 | 20 | 4 | 65.6 | 60.7 | 30 | 4 | 57.4 |
| Vehicle | 400 | 18 | 68.1 | 165 | 10 | 68.6 | 67.3 | 109 | 9 | 60.6 |
| H2 | 300 | 10 | 53.8 | 160 | 5 | 52.8 | 55.7 | 1 | 8 | 53.3 |
| Breast W | 400 | 9 | 98.0 | 154 | 8 | 98.7 | 89.0 | 171 | 6 | 96.0 |
| Pima | 468 | 8 | 66.0 | 140 | 5 | 74.7 | 56.3 | 205 | 8 | 60.0 |
| German | 500 | 24 | 66.8 | 275 | 19 | 66.8 | 64.4 | 226 | 15 | 69.2 |
| Ionosphere | 185 | 34 | 91.4 | 80 | 31 | 84.5 | 81.0 | 55 | 31 | 76.7 |
| Iris | 100 | 4 | 92.0 | 81 | 2 | 94.0 | 86.0 | 88 | 1 | 84.0 |
| Wine | 100 | 13 | 85.9 | 60 | 10 | 88.5 | 73.1 | 51 | 11 | 87.2 |
| Car | 226 | 6 | 89.6 | 139 | 3 | 82.0 | 85.2 | 117 | 3 | 85.4 |
| Austral | 400 | 14 | 76.5 | 258 | 11 | 77.6 | 74.4 | 186 | 7 | 79.3 |
| Tic-tac-toe | 600 | 9 | 76.0 | 254 | 8 | 74.6 | 74.0 | 213 | 8 | 74.1 |
| **Average** | **279.6** | **14** | **76.8** | **141.8** | **9.5** | **77.3** | **71.5** | **110.1** | **8.9** | **74.3** |

3. *Comparison with an independent procedure consisting in deleting features at first and then removing irrelevant instances*

   We compare $(FS + PS)RCG$ with a sequential procedure consisting in applying $FSRCG$ and then $PSRCG$ (called $FSRCG+PSRCG$ in the reduced table 5). Only in terms of storage reduction, two independent processes seem result in smaller storage requirement (23.6%), but accompanied by a dramatic accuracy reduction in comparison with a standard $k$NN classifier (-2.8 on average). Therefore, such a strategy would only solve part of the storage reduction problem. Using a Student paired t test, this difference is highly significant, in the detriment of $FSRCG + PSRCG$.

4. *Comparison with a single prototype selection (PSRCG) and a single feature selection (FSRCG)*

   Results in the table 5 show the interest of $(FS + PS)RCG$ in comparison with single reduction treatments. The global complexity (number of instances× number of features) in storage and calculation terms is smaller using our mixed procedure (34.4 vs 47.8 and 53.5). Moreover, its predictive accuracy is significantly better in comparison with $PSRCG$, and not distinguishable in comparison with $FSRCG$.

5. *Computational efficiency of $(FS + PS)RCG$*

   What we loose with the alternative deletion procedure, we get it back with the reduction of the instance sample size. Moreover, we can optimize the search for the $k$ nearest neighbors by using relevant algorithms such as $k$-$d$ trees. Finally, if the use of this procedure can result in greater storage reduction and then smaller computational costs during post-processing, the extra computation during learning can be very profitable.

## 5 CONCLUSION

Reducing storage requirements by selecting relevant features and prototypes seems to become a crucial problem in machine learning with the huge modern data bases. However, few works have attempted to take the reduction problem as a whole, probably because of the complexity of each subproblem. This paper brings a double novelty in this domain: it constitutes one of the first attempts to treat the storage reduction through its two degrees of freedom; it challenges the hegemony of the accuracy considered as the indisputable criterion to optimize in a selection algorithm (especially in prototype selection). We proposed in this paper a heuristic based on information theory that attempts to jointly deal with the feature and prototype selection. Our algorithm, called $(FS+PS)RCG$, is computationally suited to deal with various benchmarks. Nevertheless, while few benchmarks used by our community present more than 40 features and more than 5000 instances, some real-world domains give huge data sets on which the majority of the standard selection algorithms are incompetent. A future challenge will then consist in studying the relevance of our algorithm on more challenging data sets, which present more features and more instances. Finally, an extension of this work should analyze the nature of the neighborhood graph built on the learning set. Actually, other neighborhood structures could be applied in such a context.

Table 5: Accuracy and storage requirements for different storage reduction techniques

|  | Size | Dim | Size × Dim | Accuracy |
|---|---|---|---|---|
| 5-NN | 100.0 | 100.0 | 100.0 | 76.8 |
| PSRCG | 47.8 | 100.0 | 47.8 | 76.3 |
| FSRCG | 100.0 | 53.5 | 53.5 | 77.1 |
| FSRCG+PSRCG | 44.1 | 53.5 | 23.6 | 74.0 |
| (FS+PS)RCG | 50.7 | 67.9 | 34.4 | 77.3 |